\documentclass[runningheads]{llncs}

\usepackage{eccv}

\usepackage{eccvabbrv}

\usepackage{graphicx}
\usepackage{booktabs}
\usepackage{times}
\usepackage{soul}
\usepackage{url}
\usepackage{amsmath}
\usepackage{booktabs}
\usepackage{algorithm}
\usepackage{multirow}
\usepackage{multibib}
\usepackage{algpseudocode}
\usepackage{textcomp}
\usepackage{xcolor}
\usepackage{subcaption}
\usepackage{bm}
\usepackage{multirow}
\usepackage{amssymb}
\usepackage{array}
\usepackage{adjustbox}

\usepackage[accsupp]{axessibility}  

\usepackage[pagebackref,breaklinks,colorlinks,citecolor=eccvblue]{hyperref}

\usepackage{orcidlink}

\begin{document}

\title{FedTSA: A Cluster-based Two-Stage Aggregation Method for Model-heterogeneous Federated Learning} 

\titlerunning{FedTSA}

\author{Boyu Fan\inst{1} \and
Chenrui Wu\inst{2,3} \and
Xiang Su\inst{4} \and
Pan Hui\inst{5,1}}

\authorrunning{Fan et al.}

\institute{University of Helsinki, Helsinki, Finland 
\and
Zhejiang University, Hangzhou, China
\and
Simon Fraser University, Burnaby, Canada
\and
Norwegian University of Science and Technology, Gjøvik, Norway \and
The Hong Kong University of Science and Technology (Guangzhou), China
}

\maketitle

\begin{abstract}

Despite extensive research into data heterogeneity in federated learning (FL), system heterogeneity remains a significant yet often overlooked challenge. Traditional FL approaches typically assume homogeneous hardware resources across FL clients, implying that clients can train a global model within a comparable time frame. However, in practical FL systems, clients often have heterogeneous resources, which impacts their training capacity. This discrepancy underscores the importance of exploring model-heterogeneous FL, a paradigm allowing clients to train different models based on their resource capabilities. To address this challenge, we introduce FedTSA, a cluster-based two-stage aggregation method tailored for system heterogeneity in FL. FedTSA begins by clustering clients based on their capabilities, then performs a two-stage aggregation: conventional weight averaging for homogeneous models in Stage 1, and deep mutual learning with a diffusion model for aggregating heterogeneous models in Stage 2. Extensive experiments demonstrate that FedTSA not only outperforms the baselines but also explores various factors influencing model performance, validating FedTSA as a promising approach for model-heterogeneous FL.
  
\end{abstract}

\section{Introduction}
\label{sec:intro}

Federated learning (FL) has emerged as a promising paradigm for privacy-preserving machine learning~\cite{lim2020federated,li2021survey,10293196}. However, system heterogeneity, due to varying client capabilities, poses significant challenges for deploying FL in practice~\cite{ye2023heterogeneous,Ilhan_2023_CVPR,Liao_2023_CVPR,Zhang_2023_CVPR}. Traditional FL frameworks operate under the assumption that all clients share the same model structure~\cite{mcmahan2017communication,bonawitz2019towards,wang2020tackling}. In fact, the real-world environments include devices with different capabilities, ranging from resource-constrained devices to high-performance GPU-enabled machines. Some devices may struggle with the computational demands of the global model, leading to prolonged training times, while others might be underutilized, which can lead to inefficiencies and potential training biases~\cite{9475501,park2021sageflow}.

\begin{figure}[t]
  \centering
  \includegraphics[width=0.8\linewidth]{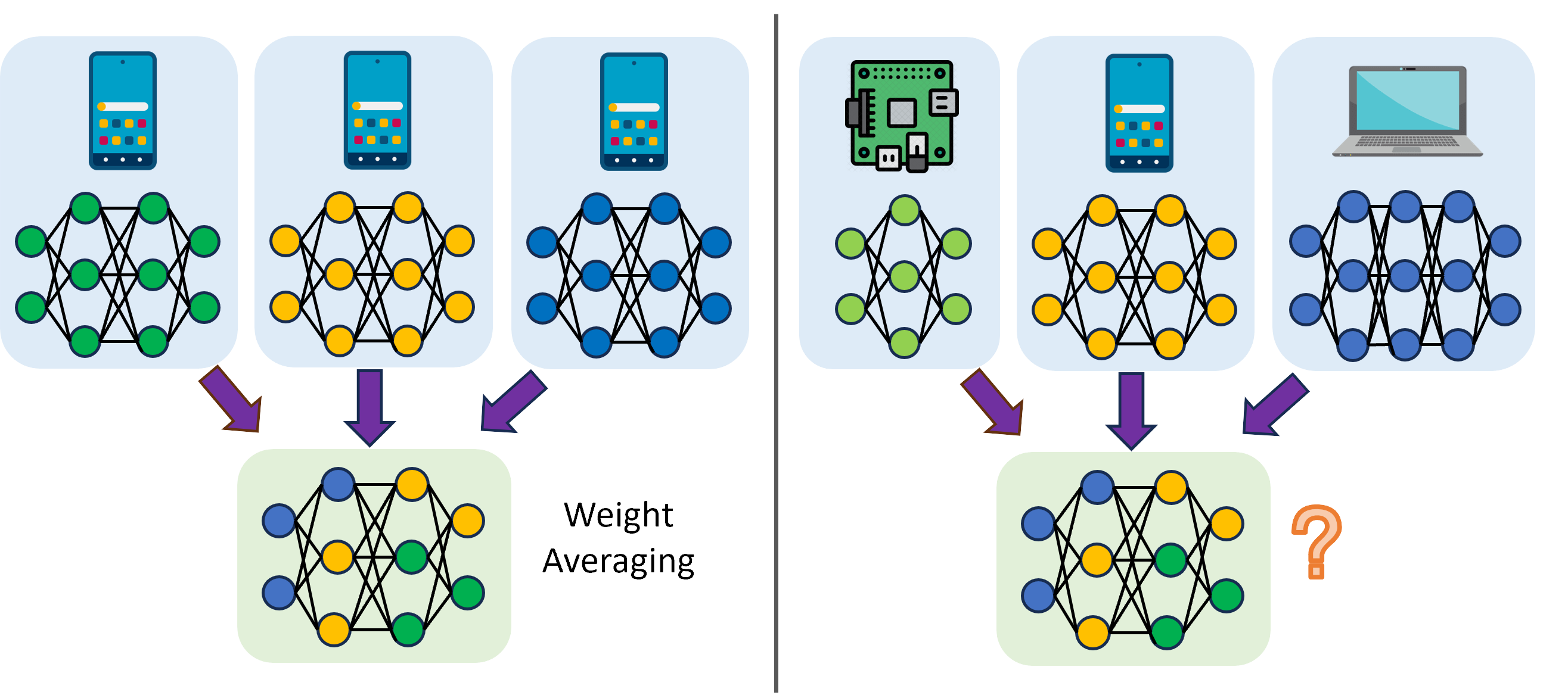}
  \caption{\textit{Left:} FL in homogeneous settings. All clients share the same global model and the server performs weight averaging to aggregate them. \textit{Right:} FL in heterogeneous settings. Due to the resource constraints, some clients can only train simple models, leading to heterogeneous models. As these models have different dimensions, how to aggregate them becomes the core challenge.}
  \label{introduction}
\end{figure}

A vanilla idea is to allow clients to adopt different model architectures tailored to their capabilities. For instance, a resource-constrained Raspberry Pi might use a two-layer convolutional neural network (CNN) model, while a GPU-equipped PC could employ a ResNet18 model~\cite{he2016deep}. When using traditional FL algorithms, such as FedProx~\cite{li2020federated}, aggregating these diverse architectures becomes challenging due to the varying dimensions of model parameters, as illustrated in Figure~\ref{introduction}. This discrepancy makes the weight averaging on the server infeasible. Knowledge distillation (KD)-based approaches have been proposed to mitigate this challenge by only aggregating and aligning with the global knowledge~\cite{li2019fedmd,Gong_Sharma_Karanam_Wu_Chen_Doermann_Innanje_2022}. However, they often require a public dataset on the client side, raising privacy and resource concerns~\cite{10547356}. Data-free KD-based methods have emerged as potential solutions to mitigate this limitation~\cite{zhu2021data,lin2020ensemble}, but they come with substantial communication costs and the issue of low-quality generated data. Another strategy is partial training, where each client trains a subnetwork of the global model~\cite{diao2021heterofl,alam2022fedrolex} and only the overlapping subnetworks will be aggregated. However, a critical issue in partial training is the non-overlapping model parameters, which may contain important information but never get updated.

In this paper, we propose FedTSA, a cluster-based two-stage aggregation method for model-heterogeneous FL. For any given FL training task, FedTSA starts with clustering clients according to their hardware resources using kernel density estimation (KDE)~\cite{kim2012robust}. Afterwards, FedTSA assigns varying pruning rates to different clusters, leading to heterogeneous models across clusters. Within each cluster, the homogeneous clients conduct local updates and perform model aggregation in conventional FL training styles, a process we refer to as Stage 1 aggregation. To aggregate heterogeneous models from Stage 1, we employ deep mutual learning (DML)~\cite{Zhang_2018_CVPR} on the server based on the data generated by a diffusion model~\cite{NEURIPS2020_4c5bcfec}. This phase is refer to as Stage 2 aggregation. The employment of the diffusion model not only obviates the need for crafting public datasets but also fosters the generation of superior-quality data, enhancing DML efficacy. Moreover, unlike FedGen~\cite{zhu2021data}, computation-intensive data generation tasks are offloaded to the server side, with clients only sending prompts to guide this process, which ensures a high degree of flexibility in training tasks.

Our extensive experiments on computer vision (CV) datasets demonstrate FedTSA's effectiveness, outperforming other state-of-the-art model-heterogeneous FL approaches. We also provide an in-depth analysis of various factors influencing FedTSA, offering valuable insights for future model-heterogeneous FL research.
Our contributions are threefold:
  1) We introduce the novel FedTSA framework to enable model-heterogeneous FL. FedTSA first clusters clients according to their processing capabilities using KDE, then conducts a two-stage aggregation to aggregate heterogeneous models.
  2) FedTSA innovatively leverages a diffusion model to generate data, thereby addressing the challenges of constructing a public dataset. This approach not only enhances data privacy but also significantly boosts the framework's adaptability to various training tasks.
    3) Extensive experimental evaluations on three CV datasets demonstrate FedTSA's superior performance over existing baselines. Moreover, our investigations into influential factors present critical insights, further demonstrating FedTSA's effectiveness.

\section{Related work}
\label{related_work}

\textbf{Model-heterogeneous FL.} Different from the conventional FL paradigm~\cite{mcmahan2017communication,10092911,li2020federated,10219725}, model-heterogeneous FL~\cite{fan2023modelheterogeneous} allows clients to train models of varying complexities based on their processing abilities, thus mitigating the system heterogeneity issue. Inspired by the idea of KD~\cite{hinton2015distilling,phuong2019towards}, \cite{li2019fedmd} proposes FedMD, a logit-based FL algorithm. Instead of transferring model parameters, FedMD employs logits as the communicative data medium and trains the global model by aggregating and approaching these logits. However, FedMD relies on a public dataset to align the input data, leading to privacy concerns and extra communication costs. This problem exists in several similar works~\cite{chang2019cronus,jeong2018communication}. Subsequent works such as FedDF~\cite{lin2020ensemble} and FedGen~\cite{zhu2021data} address these limitations by either leveraging unlabeled data or training a generator to produce synthetic data to achieve data-free KD. However, FedDF requires prior knowledge to choose proper unlabeled data, while FedGen suffers from high computation cost as the generators are deployed and worked in each client. Moreover, the quality of unlabeled data or data generated by generative adversarial networks (GAN)~\cite{goodfellow2014generative} cannot be guaranteed, potentially leading to a decrease in model performance. Based on the idea of class prototype~\cite{snell2017prototypical}, FedProto~\cite{fedproto} aggregates heterogeneous models by exchanging and aligning prototypes. In addition, HeteroFL~\cite{diao2021heterofl} achieves model-heterogeneous FL by only aggregating the overlapping submodel parts, thus enabling clients to train submodels with different sizes. Despite existing methods adopt different strategies to achieve model-heterogeneous FL, none of them illustrate how to allocate different models to the clients.

\textbf{Diffusion Models.} With large-scale training data and model parameters, various text-to-image generative models are now capable of creating high-quality images~\cite{qiao2019mirrorgan,dhariwal2021diffusion,ruiz2023dreambooth}. Among them, diffusion models~\cite{NEURIPS2020_4c5bcfec} have achieved significant success in the image synthesis domain and AI-generated content (AIGC) fields, such as the popular Stable Diffusion~\cite{Rombach_2022_CVPR} and DALL-E2~\cite{ramesh2022hierarchical}. A typical diffusion model includes a forward diffusion stage and a reverse diffusion stage~\cite{10081412}. Given a textual prompt, the diffusion model is able to convert Gaussian noise into a text-compliant image through an iterative denoising step. Despite diffusion models showcasing excellent performance, only a few works have leveraged them in FL. \cite{yang2023exploring} and~\cite{zhao2023federated} combine diffusion models with FL to mitigate non-IID problems, while~\cite{tun2023federated} applies FL to the training process to address privacy issues. 

FedTSA is the first work that leverages the advantages of diffusion models to achieve model-heterogeneous FL. Compared to other methods that either use a well-crafted public dataset or train a generator on the client side, employing a diffusion model not only improves the quality of the generated data but also offers greater flexibility in changing training tasks and better privacy protection.

\section{Motivation Study}
\label{motivation}
We motivate this study by presenting the impact of system heterogeneity through a small-scale experiment. We establish a heterogeneous system comprising four types of devices, including three Raspberry Pi 4B units, three LattePanda 2 Alpha 864s~\cite{lattepanda}, two PCs without GPU, and one PC equipped with an NVIDIA 3080Ti GPU. These devices range from embedded systems to PCs with GPUs, effectively presenting the system heterogeneity prevalent in real-world scenarios. The core specifications of each device type are detailed in Table~\ref{devicespec}. Given the limited resources of the Raspberry Pi, we select a FL training task using the MNIST dataset and a CNN model. This model includes two convolutional layers with max pooling, followed by two fully connected layers. Each device serves as a FL client, with an additional machine serving as the central server for aggregation. We employ FedAvg~\cite{mcmahan2017communication} in this study.

\begin{table}[h]
\begin{minipage}{0.45\textwidth}
\centering
\caption{Device Core Specifications.}
\label{devicespec}
\begin{tabular}{c|ccc}
\toprule
 Device         &  CPU & Memory & GPU  \\ 
\midrule
Rasp Pi     & A72, 1.8GHz & 8GB &       N/A         \\ 
LattePanda  &     M3, 3.4GHz       &       8GB          &  N/A             \\ 
PC       &     i7, 2.2GHz       &        16GB         &  N/A              \\ 
PC with GPU       &     i7, 2.4GHz       &        32GB         &  3080Ti               \\ 
\bottomrule
\end{tabular}
\end{minipage}\hfill
\begin{minipage}{0.45\textwidth}
\centering
\caption{Wall-clock convergence times in different settings.}
\label{motivation}
\begin{tabular}{c|c}
\hline
\textbf{Setting} & \textbf{Time (s)} \\ \hline
All devices involved                &    2731                          \\ \hline
No Rasp Pi                 &    1428 (-47\%)                       \\ \hline
No Rasp Pi \& LattePanda                 &      210 (-92\%)                    \\ \hline
Only PC with GPU                 &      143 (-95\%)                   \\ \hline
\end{tabular}
\end{minipage}
\end{table}

We design four experiments to assess the impact of system heterogeneity. The settings and the wall-clock convergence times are detailed in Table~\ref{motivation}. We can observe that when all devices participate in training, the convergence speed is extremely slow, exceeding 2700 seconds. This is attributed to tail latency~\cite{li2014tales} caused by the existence of Raspberry Pi devices, which delay the process even after other devices have completed their training. The convergence times for the remaining three experiments are reduced to 1428, 210, and 143 seconds, respectively, showing decreases of 47\%, 92\% and 95\% compared to the initial scenario. These results show the idle times of faster devices, highlighting the significant impact of system heterogeneity on FL training efficiency. 

Some existing works mitigate this issue by focusing on communication strategies, e.g., by designing asynchronous or semi-asynchronous FL algorithms to improve training efficiency~\cite{chen2020asynchronous,wu2020safa,10.1145/3639825}. However, such methods either introduce excessive additional communication overhead or suffer from stale models. In this paper, we address this issue fundamentally by facilitating model-heterogeneous training. Specifically, we allow different devices to select models that are well-suited to their system resources, thereby effectively addressing the issue of prolonged waiting times due to insufficient resources.

\section{Methodology}
\label{methodology}

\subsection{Problem Definition}
In this work, we consider a system heterogeneity setting in FL. Suppose we have $n$ clients with their private local dataset $D_{i}$ from distribution $P_{i}(x, y)$, with $x$ and $y$ representing the input data and their respective labels. Unlike most works that assume clients have similar hardware resources, we consider the presence of heterogeneous resource distributions among clients, denoted as $R_{1}, R_{2}, ..., R_{n}$. Due to the challenge of heterogeneity, a standardized global model $W_{g}$ loses its applicability, as clients with limited resources cannot afford the computational costs if $W_{g}$ is complex. Therefore, the problem becomes how to enable clients to deploy different models $W_{1}, W_{2}, ..., W_{n}$ according to their respective resources $R_{1}, R_{2}, ..., R_{n}$, while still achieve FL training on the server side. Similar to conventional FL approaches, for the $i^{th}$ client, the training procedure is to minimize the loss as follows:
\begin{equation}
\arg \underset{W_{i}}{\min}  \sum_{i=1}^{n} \frac{\left | D_{i} \right | }{N}\mathcal\ell_{ce}(f(W_i; x), y),
\label{cluster_cost}
\end{equation}
where $N$ is the total number of the training data over all clients, $f(W_i,x)$ denotes the output of model $W_{i}$ give the input data $x$, and $\mathcal\ell_{ce}$ is the cross entropy-loss. However, the varied architectures of models $W_{i}$ imply that the model parameters from clients are incompatible with elementary weight averaging, since they have different dimensions. Therefore, a new strategy is required to aggregate these model parameters on the server side to achieve global update. To tackle this dilemma, we propose a two-stage aggregation framework FedTSA, where clients with similar resources conduct conventional weight averaging within a cluster, while clients with significantly different resources contribute to each other by DML outside the cluster. 

\subsection{Two-Stage Aggregation Framework}
\begin{figure*}[h]
\centering
\includegraphics[width=0.95\textwidth]{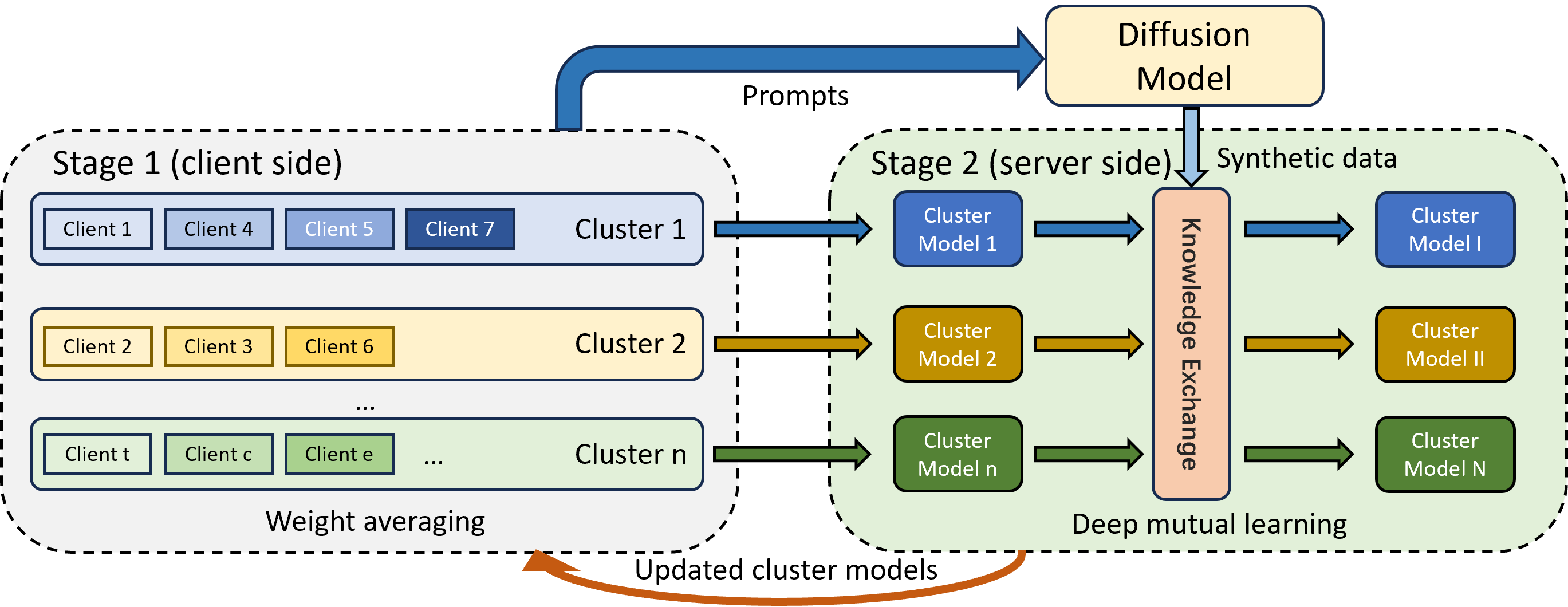}
\caption{FedTSA framework with two-stage aggregation.}
\label{framework}
\end{figure*}

Figure~\ref{framework} presents the FedTSA framework, highlighting two stages of aggregation, i.e., in-cluster weight averaging aggregation (Stage 1) and server-side DML aggregation (Stage 2). In Stage 1, we harness a resource-driven clustering methodology to categorize clients, leading to $n$ distinct clusters. Each cluster receives a unique pruning rate, creating models of varying complexities. Then, clients conduct local updates, but only updates from the same cluster perform weight aggregation, as they share an identical model architecture and thus possess equivalent parameter dimensions. These cluster-based models are sent to the server for Stage 2 aggregation.

Stage 2 in Figure~\ref{framework} depicts how models with different architectures achieve aggregation with the help of DML and a diffusion model. Here, data generated from the diffusion model enables knowledge transfer between heterogeneous models. Specifically, each model initially takes these data as inputs to generate logits, which represent the output vectors of the neural network before applying the softmax function. Afterwards, the models perform gradient descent using the same generated data with Kullback Leibler (KL) divergence loss. Contrary to conventional KD-based FL methods, in our setting, no global model exists on the server side. Therefore, KL divergence loss is calculated by comparing each set of logits with a consensus. In this context, the consensus is defined as the average of all logits generated by different models. After the DML process, knowledge is interchanged among models, finishing Stage 2 aggregation. The updated models are sent back to their respective clusters for further updates. Algorithm~\ref{alg_fedtsa} outlines the FedTSA learning process, where the LocalUpdate function is the same as FedAvg.

The diffusion model serves as a fundamental component in this framework, offering three advantages over other KD-based FL methods. First, it eliminates the need for a well-prepared public dataset deployed on server and client sides, thereby reducing client-side computational costs and manpower. Second, it produces higher quality data than GAN-based methods~\cite{dhariwal2021diffusion}, improving the quality of logits and benefiting the DML process. Third, by simply altering the prompts, clients are enabled to seamlessly transition between different training tasks.

\begin{algorithm}[!t]
    \caption{FedTSA}
    \label{alg_fedtsa} 
    \begin{algorithmic}
        \State \textbf{Input:} $m$ clients, prompt $P$ 
        \State Server initializes the proxy task to the clients. 
        \State Server collects the running time \bm{$T$} from all clients. 
            \State clusters$\leftarrow$\textbf{Clustering($T$)} \Comment{Clustering is implemented according to Sec~\ref{clustering_strategy}}
            \For{$t=0, 1, \ldots, T-1 $}
            \For{cluster $c\in$ clusters in parallel} 
            \For{client $i\in$ cluster $c$ in parallel} 
            \State $w_{c}^{i} \leftarrow$ LocalUpdate$(i, w_{c}^{i})$ 
            \EndFor
            \State $w_{c}^{t} = \frac{1}{\left | c \right | } {\textstyle \sum_{k\in c}^{}w_{k}} $ \Comment{Models within a cluster perform simple weight averaging}
            
            \EndFor
            \State $w_{c}^{t+1}\leftarrow$\textbf{DML}($w_{c}^{t}$) \Comment{Models from different clusters perform DML to be aggregated}
            \EndFor

            \Statex
            \Function{DML}{$\bm{w_c}$}
            \State Diffusion model generate data $D_i$ according to $P$. 
            \For{each global epoch}
            \For{batch $\left(x_{i}, y_{i}\right) \in D_i$}
            \State $p_i$ $\leftarrow$ Eq.\ref{eq:logits} \Comment{Calculate the soft prediction of model $w$}
            \State $L_{kl} \leftarrow$ Eq.\ref{eq:average_logit} \& Eq.~\ref{eq:kl_loss} \Comment{Calculate the KL loss with the averaged logits}
            \State $w_c^i=w_c^i-\eta \nabla L_{kl}$ \Comment{Update the cluster-based model}
            \EndFor
            \EndFor
            \State
            \Return $\bm{w_c}$ \Comment{Return the updated cluster-based model in cluster c}
            
            \EndFunction

    \end{algorithmic} 
\end{algorithm}

\subsection{Resource-oriented Clustering with Pruned Models}
\label{clustering_strategy}
In our FedTSA framework, clustering serves as the initial step, grouping clients according to hardware resources to enhance efficient training and reduce tail latency. Unlike most existing works that group clients to address data heterogeneity~\cite{ouyang2021clusterfl,Caldarola_2021_CVPR}, our focus is on ensuring resource homogeneity within clusters. While one could categorize resources by hardware like CPU and memory, solely using these attributes is insufficient. Environmental factors, such as temperature shifts and voltage variations, can impact device performance. Moreover, the network condition plays a fundamental role, especially given the multiple communication rounds inherent to the FL paradigm. 

We propose to employ a proxy task to acquire the training duration of each client, which will act as a key attribute for clustering. This duration serves as a reflection of the run-time operational efficiency of a given system. For the proxy task, the server dispatches a one-hidden-layer multilayer perceptron (MLP) along with a scaled-down MNIST dataset to the clients. Once the clients receive the model and data, they start the training process and simultaneously record the training durations. These logged durations are then sent back to the server, serving as the essential attribute for the clustering procedure. It is noteworthy that this proxy task can be any lightweight training task that can run on all clients participating in FL.

Our clustering method is based on KDE~\cite{kim2012robust}, which is a non-parametric approach for estimating the density of a distribution. Specifically, given a set of $n$ points ($x_{1}, x_{2}, ...,\\ x_{n}$) which refer to the training durations from the proxy task and reflect the system capacities, the kernel density estimator $\hat{f} (x)$ is given by
\begin{equation}
\hat{f} (x) = \frac{1}{nh}\sum_{i=1}^{n}K(\frac{x-x_{i}}{h}),
\label{cal_alpha}
\end{equation}
where $h$ is the smoothing parameter named bandwidth and $K$ is a kernel function that meets the following condition:
\begin{equation}
\int_{-\infty }^{\infty } K(x)dx=1.
\end{equation}
Given the heterogeneity of device resources in the wild is roughly presented as a Gaussian distribution, we leverage the Gaussian kernel for the estimator given by:
\begin{equation}
K(x)=\frac{1}{\sqrt{2\pi }}\exp (-\frac{x^{2} }{2}).
\end{equation}
With the estimated density function, we can easily acquire a series of local maximum and minimum points, which can naturally serve as the boundaries of different clusters. After clustering, we compute the average processing duration of the clients within each cluster and subsequently map them onto a pruning rate distribution. The cluster with the minimal mean processing duration is assigned a pruning rate of 1.0, indicating an unpruned model. Other clusters' rates can be calculated by $ratio_i=\frac{t_f}{t_i}$, where $t_f$ is the fastest during and $t_i$ is the current one.

Neural network pruning has gained significant attention as a technique for decreasing complexity~\cite{MLSYS2020_6c44dc73,Wang_2021_CVPR,pmlr-v119-tan20a}. Leveraging pruning, the amount of computation required to train neural networks can be greatly reduced. Based on this principle, we can create models with diverse computational requirements to cater to different client resources. The pruning rate $p \in (0,1]$ serves as a parameter to modify the architectures of various neural networks. For instance, a fully connected neural network (FCNN) can be pruned by adjusting the number of neurons in each layer, while a CNN can vary the number of input/output channels. For a given $p$ value, the weights of a pruned network can be represented as $W^{p}$. This allows us to form different model-heterogeneity levels. Each cluster has a unique $p$ value, so clients within a cluster share the same architecture, termed the cluster-based model. Details on server-side aggregation of these models are provided in the subsequent section.

\subsection{DML with Diffusion Model}
In this section, we present the process of server-side DML with the help of a diffusion model. Denoising diffusion probabilistic models (DDPM)~\cite{NEURIPS2020_4c5bcfec} have been developed as probabilistic models, aiming at learning the underlying data distribution $p(x)$. This is achieved through a gradual denoising process applied to a normally distributed variable, which corresponds to the inverse process of a fixed Markov Chain with a length of $\mathcal{T}$,
\begin{equation}
    x_t=\sqrt{\alpha_t}x_{t-1}+\sqrt{1-\alpha_t}\epsilon_t, t\in \{1,2,\cdots,\mathcal{T}\},
\end{equation}
where sequence $\{x_t\}$ changes from $x_1$ to $x_T$ sampled from a Gaussian distribution $\mathcal{N }(0, I)$. At each step, Gaussian noise $\epsilon_t$ is added, drawn from $\mathcal{N }(0, I)$. $\{ \alpha_t \}_{t=1}^{\mathcal{T}}$ is pre-defined constants as the vanilla DDPM. The denoising process is designed to be the inverse of the diffusion process, wherein the Gaussian noise $x_{\mathcal{T}} \sim \mathcal{N }(0, I)$ is progressively restored back to its original state $x_1$ through denoising steps, for $t = \mathcal{T},\cdots, 1.$

To enhance efficiency, FedTSA avoids the long training process that is usually required for diffusion models. We simply develop a prompt pool denoted as $\mathcal{Y}$ and leverage a pre-trained Stable Diffusion v1-4 Model~\cite{Rombach_2022_CVPR} for inference. By allowing clients to upload prompts to the server's prompt pool, we have effectively mitigated concerns about data distribution and potential privacy issues. Moreover, before uploading textual prompts, the implementation of differential privacy techniques~\cite{sun2020ldp} can be employed, whereby the inclusion of irrelevant text further safeguards the privacy of the system. 

In the diffusion model inference stage, when provided with a text prompt $y\in \mathcal{Y}$ sent by the clients, the denoising U-Net~\cite{ronneberger2015u} $\epsilon_\theta (\cdot)$ generates predictions for the data sample $x$, taking into consideration the conditioning on the text $y$. These generated data $\mathcal{X}=\{x_i\}_{i=1}^N$ is used to conduct knowledge exchange between heterogeneous cluster-based models. Inspired by DML, we observe that mutual guidance between models enhances their generalization abilities. Unlike conventional KD, which may cause a student model to overfit to the teacher model’s outputs, DML introduces increased robustness and diversity. When applied to collaborative training across clusters, DML optimally harnesses inter-cluster knowledge, enabling heterogeneous cluster-based models to communicate. The soft prediction of a model $w$ for a given sample $x_i \in \mathcal{X}$ for a class $c \in \{1,2,\cdots,C\}$ is denoted as:
\begin{equation}
\label{eq:logits}
    p_i^c=\frac{\text{exp}(z_i^c/T)}{\sum_{m=1 }^{C}\text{exp}(z_i^m/T)},
\end{equation}
where $z_i^c$ is the logits, denoting the $i$-th element in the model classifier’s outputs $\mathcal{P}=\{ z_i^c \}_{c=1}^{C}$ for a given sample $x_i$. $T$ is a temperature parameter to adjust the impact of logits. Assuming the output logits of a cluster-based model $W_r$ on the generated data are $z_r$, we compute the global knowledge for cluster-based models with the average logits during the DML process, formulated as:
\begin{equation}
\label{eq:average_logit}
z_{avg}=\frac{1}{M} \sum_{r=1}^{M} z_r,
\end{equation}
where $M$ is the total number of clusters. To evaluate the divergence between two predictions, we apply the KL divergence:
\begin{equation}
\label{eq:kl_loss}
D_{KL}(p_{2}\Vert p_{1})=\sum_{i=1}^{N}\sum_{c=1}^{C}p_{2}^{c}(x_{i})\log\frac{p_{2}^{c}(x_{i})}{p_{1}^{c}(x_{i})}.
\end{equation}
The last step is calculating the KL divergence between the logits of each cluster-based model and the global knowledge, acting as a loss function to guide the model update. After each cluster gets the updated cluster-based model from the server, the clients start the local training for the next round with the new model and repeat the Stage 1 and Stage 2 aggregation until the model converges.

\section{Evaluation}
\label{experiments}

\subsection{Experimental Setup}
\textbf{Datasets and Models.} We evaluate the performance of FedTSA using three benchmark datasets, i.e., CIFAR-10, CIFAR-100~\cite{krizhevsky2009learning}, and Tiny-ImageNet~\cite{chrabaszcz2017downsampled}. We consider both IID and non-IID data distributions in the experiments. In the IID setting, data is evenly distributed to each client in terms of both quantity and class distribution. In the non-IID settings, we employ the Dirichlet distribution $Dir(\alpha)$~\cite{li2022federated} to allocate the dataset among the clients, where a smaller $\alpha$ indicates higher data heterogeneity. We set the $\alpha$ to 0.3 and 0.6 to introduce different data heterogeneity levels. We use the ResNet18 as the model in the experiments. To incorporate model heterogeneity, we adjust the model complexity by varying the hidden sizes of the residual blocks with a pruning rate, where a smaller rate represents a simpler model. 

\textbf{Hyper-parameters.} In the experiments, the hyper-parameters are set as follows. For the general FL training parameters, we set the local update batch size $B$ to 100, the number of local epoch $E$ to 100, the number of users $K$ to 20, the learning rate to 0.03, and the training round to 100. For FedTSA's specific hyper-parameters, we set the temperature $T$ to 5, the number of global epochs to 1, and the DML loss function to KL-only. In terms of the prompts for the diffusion model, we use the format of ``A photo of real $\ast$'', where $\ast$ can be ``bird'', ``cat'' and other objects according to the training tasks. We also explore the impact of these specific hyper-parameters on model performance. 

\textbf{Baselines and Implementation.} Given that the primary contribution of FedTSA is to achieve promising model-heterogeneous FL, we select four FL approaches that allow clients to use different model architectures, i.e., FedProto~\cite{fedproto}, FedMD~\cite{li2019fedmd}, FedDF~\cite{lin2020ensemble} and HeteroFL~\cite{diao2021heterofl}. We also select the vanilla FL algorithm FedAvg~\cite{mcmahan2017communication} and the improved version FedProx~\cite{li2020federated} to underscore the significance of enabling model heterogeneity. For model-homogeneous algorithms, FedAvg and FedProx, we present the results where clients have complete models and pruned models ($p$=0.6), representing upper and lower accuracy bounds, respectively. For model-heterogeneous algorithms, FedProto, FedMD, FedDF, HeteroFL and FedTSA, we employ three clusters with pruning rates of 1.0, 0.8, and 0.6. These rates correspond to client distributions of 20\%, 40\%, and 40\%, respectively. The experimental results are based on the average of five runs. All experiments are implemented in PyTorch and executed on a server machine with an NVIDIA A100 GPU.

\subsection{Performance Evaluation and Analysis}
\textbf{Performance Comparison.} From the results presented in Table~\ref{acc_comp}, we can observe that FedTSA outperforms other FL baselines across both IID and non-IID settings, highlighting its promising performance and robustness to heterogeneous data. An exception is noted in the IID setting of CIFAR-10 dataset, where FedTSA slightly lags behind the upper bound performance of FedAvg. However, we should notice that the upper bounds for FedAvg and FedProx are based on an idealized scenario where all clients have unpruned models. Given this context, FedTSA still maintains a dominant performance over most upper bounds across both IID and non-IID settings. This superior performance is attributed to FedTSA's innovative strategy. Contrasting with algorithms that conventionally replace the global model with a local model for round updates, FedTSA fosters inter-client learning through DML. Specifically, FedTSA utilizes the KL divergence loss to learn the average logits from other client models, thereby effectively obviating the need to rely on a singular global model. Consequently, each model not only gains knowledge from its counterparts but also sidesteps potential information loss that may arise from direct replacement. Besides, we can notice that FedProto presents the lowest accuracy, revealing that prototype-based FL methods do not perform as well as methods based on knowledge distillation and local training in the presence of model heterogeneity. This is because simpler models produce lower-quality prototypes, which in turn degrades the quality of the aggregated global prototypes, ultimately leading to poor model training outcomes.

\renewcommand{\arraystretch}{1.2}
\begin{table*}[h]
\caption{Average test accuracy (\%) on three datasets under different degrees of data heterogeneity.}
\label{acc_comp}
\centering
\begin{tabular}{c|c|ccccccc}
\hline
Dataset                   & IID/non-IID & FedAvg                & FedProx       & FedProto & FedMD & FedDF & HeteroFL & FedTSA \\ \hline
\multirow{3}{*}{CIFAR-10}  & IID         & 84.23-\textbf{90.86}  & 83.85-86.47   & 59.6     & 74.26        & 79.01 & 87.61    & 89.28  \\
                          & Dir(0.6)    & 81.54-83.55           & 82.31-84.26   & 49.51    & 71.33        & 76.98 & 85.09    & \textbf{86.72}  \\
                          & Dir(0.3)    & 77.25-80.91           & 80.71-82.39   & 41.33    & 70.03        & 75.12 & 82.77    & \textbf{84.61}  \\ \hline
\multirow{3}{*}{CIFAR-100} & IID         & 47.69-50.73           & 48.27-51.96   & 36.41    & 35.24        & 39.71 & 58.92    & \textbf{60.56}  \\
                          & Dir(0.6)    & 46.94-50.12           & 47.95-51.23   & 31.07    & 30.33        & 35.98 & 56.04    & \textbf{60.02}  \\
                          & Dir(0.3)    & 45.28-48.31           & 46.24-49.02   & 28.46    & 29.78        & 33.25 & 53.37    & \textbf{59.57}  \\ \hline
\multirow{3}{*}{Tiny-ImageNet} & IID    & 29.88-31.04           & 30.12-31.87   & 18.55    & 19.23        & 22.76 & 34.57    & \textbf{36.28}  \\
                          & Dir(0.6)    & 27.31-29.53           & 28.05-30.07   & 17.84    & 18.65        & 21.14 & 33.82    & \textbf{35.62}  \\
                          & Dir(0.3)    & 26.59-27.49           & 27.23-29.38   & 17.03    & 17.49        & 20.68 & 32.73    & \textbf{35.07}  \\ \hline
\end{tabular}
\end{table*}

\textbf{Effect of Different Degrees of Model Heterogeneity.} We evaluate the performance of FedTSA across varying degrees of model heterogeneity, comparing it with other model-heterogeneous FL baselines. To introduce model heterogeneity, we continue to use the pruning rate and keep the number of clients to 20. We select CIFAR-10 as the dataset and configure the data distribution to be IID. The specific settings for heterogeneity and the corresponding experimental results are detailed in Table~\ref{acc_heterogeneity}.

\begin{table*}[h]
\centering
\caption{Average test accuracy (\%) on CIFAR-10 under different degrees of system heterogeneity.}
\label{acc_heterogeneity}
\begin{tabular}{c|ccccc}
\hline
Heterogeneity degrees           & FedProto & FedMD & FedDF & HeteroFL & FedTSA \\ \hline
{[}1.0{]}                     &   74.31 & 85.23 & 87.62 &   91.13  &  \textbf{91.56} \\ \hline
{[}0.5, 1.0{]}                &   62.13  & 78.11 &  82.43  &  83.30  & \textbf{89.61}  \\ \hline
{[}0.6, 0.8, 1.0{]}           &    59.60  & 74.26 &  79.01 &  82.61 & \textbf{89.28} \\ \hline
{[}0.1, 0.4, 0.7, 1.0{]}      &   50.03 & 64.31 &  70.55  &  68.21  &  \textbf{82.37}  \\ \hline
{[}0.1, 0.3, 0.5, 0.7, 1.0{]} &  48.87  & 63.16 & 68.83 & 67.79 & \textbf{80.09} \\ \hline
\end{tabular}
\end{table*}

We can observe that FedTSA outperforms the baseline methods across all five degrees of heterogeneity. This performance illustrates FedTSA's robust capability in knowledge exchange within heterogeneous model environments. Notably, the accuracy of HeteroFL decreases significantly when the heterogeneity degrees shift from [0.6, 0.8, 1.0] to [0.1, 0.4, 0.7, 1.0]. The decline can be attributed to HeteroFL's reliance on a partial training mechanism, where only the common parts of heterogeneous models are aggregated. Therefore, if the pruning rate is excessively low, e.g., 0.1, the resulting model becomes overly simplistic, failing to effectively learn data patterns and thus contributing minimally to the global model. In contrast, FedTSA gains contributions from heterogeneous models through DML, where models with larger pruning rates can guide those with smaller ones, just like KD. That is the reason why FedTSA does not suffer from a significant decrease in accuracy. By effectively harnessing the strengths of models across a spectrum of heterogeneity, FedTSA demonstrates superior adaptability and efficiency in learning from diverse model architectures, thereby enhancing overall performance in model-heterogeneous FL scenarios.

\begin{figure}[h]
  \centering
  \begin{subfigure}[b]{0.35\linewidth}
    \includegraphics[width=\linewidth]{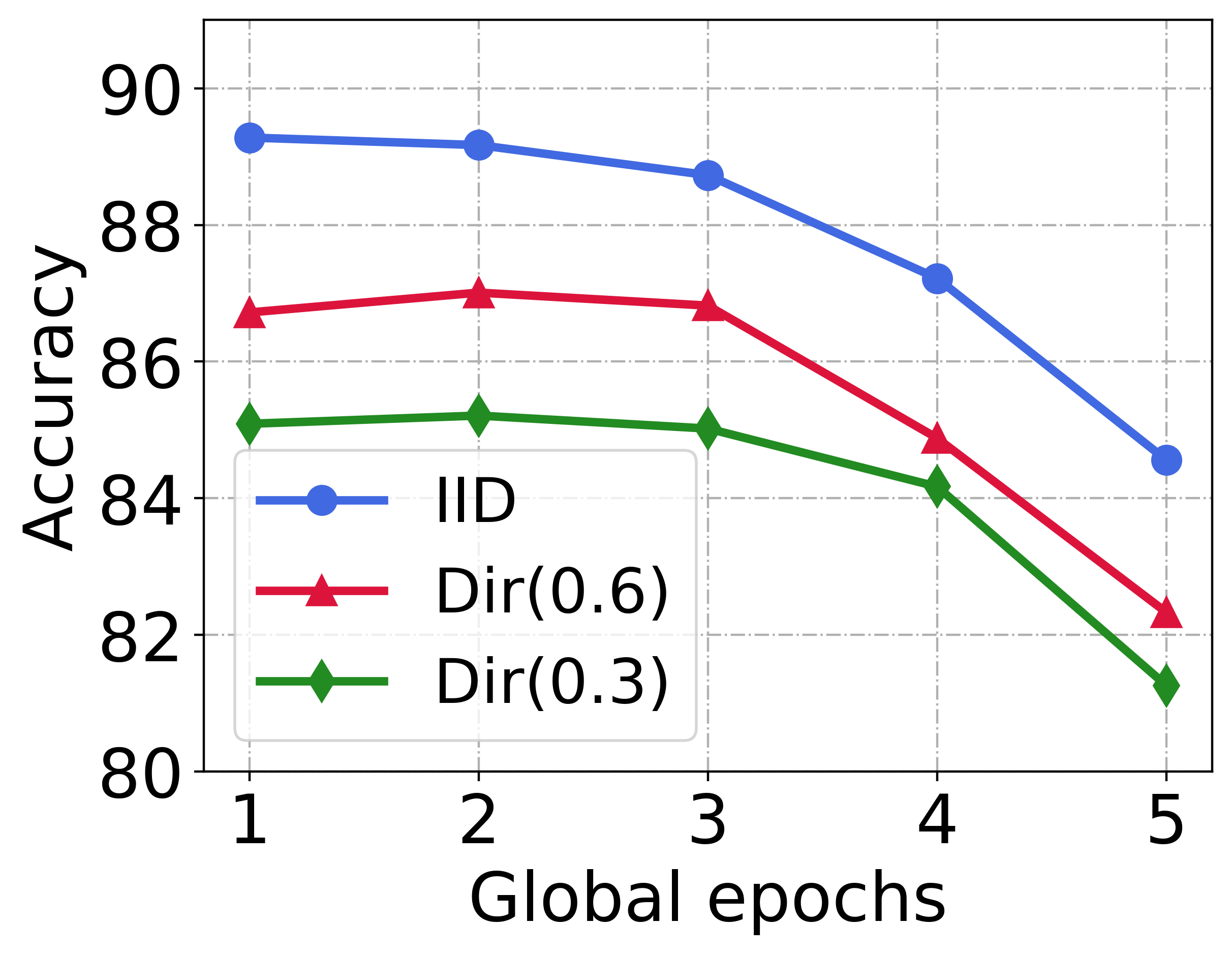}
    \caption{CIFAR-10}
    \label{fig:subfig1}
  \end{subfigure}
  \begin{subfigure}[b]{0.35\linewidth}
    \includegraphics[width=\linewidth]{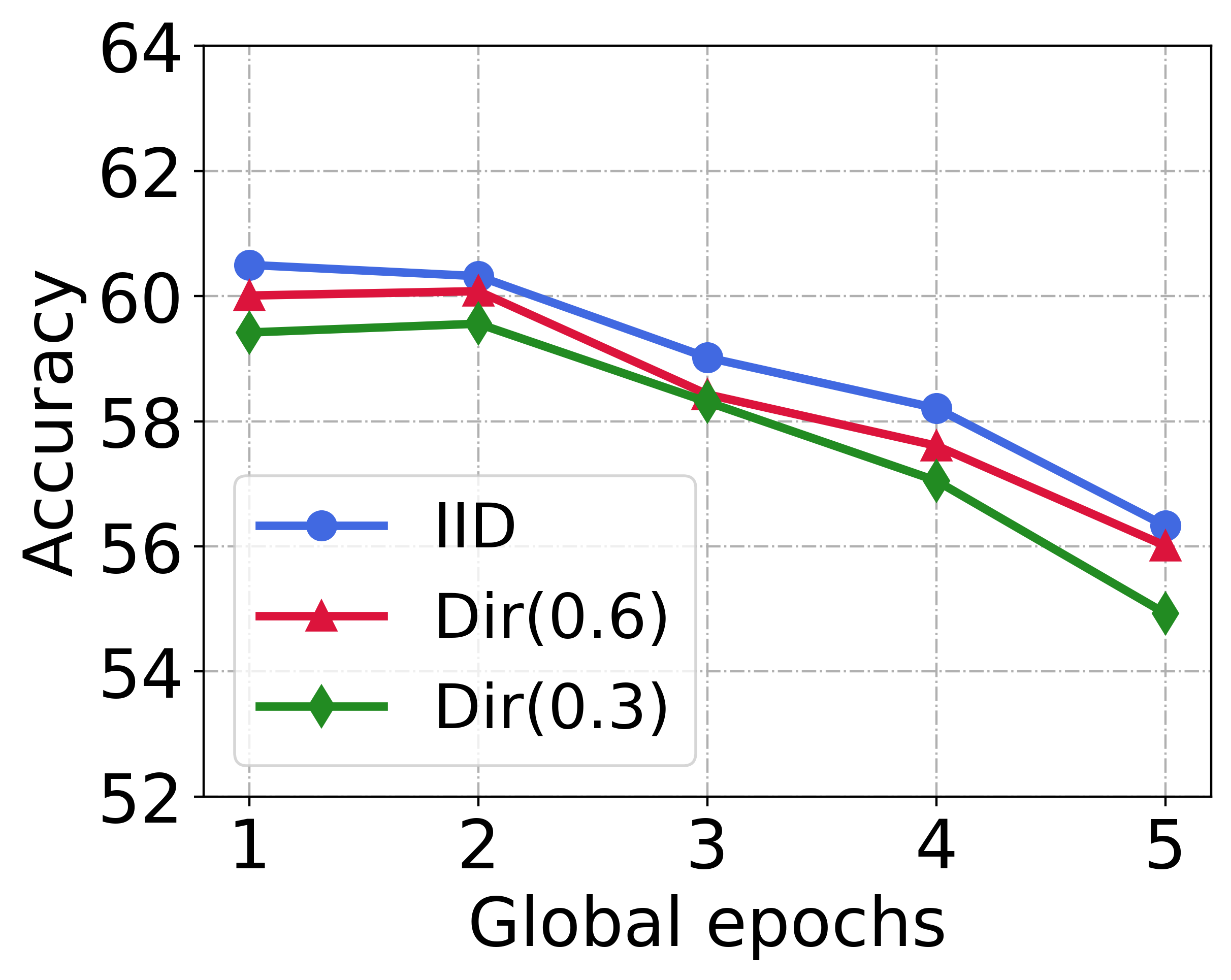}
    \caption{CIFAR-100}
    \label{fig:subfig2}
  \end{subfigure}
  \caption{Test accuracy (\%) with different global epochs in the DML process.}
  \label{fig:globalepoch}
\end{figure}

\textbf{Effect of Global Epochs.} 
Global epochs represent the epochs on generated data during the DML process. As illustrated in Figures~\ref{fig:subfig1} and~\ref{fig:subfig2}, the number of global epochs significantly influences model accuracy for both CIFAR-10 and CIFAR-100 datasets. Contrary to intuition, more global epochs reduce training efficacy. This occurs because the DML process relies on diffusion model-generated data instead of real data from various clients. The primary purpose of DML is to facilitate knowledge exchange among heterogeneous models, utilizing synthetic data as a medium for this interaction. However, too many global epochs cause models to lose their unique dataset features due to the continuous blending of average knowledge via KL divergence loss. Therefore, we limit global epochs to a single iteration for optimal DML performance.

\begin{figure}[h]
  \centering
  \begin{minipage}{0.45\textwidth}
    \centering
    \includegraphics[width=\linewidth]{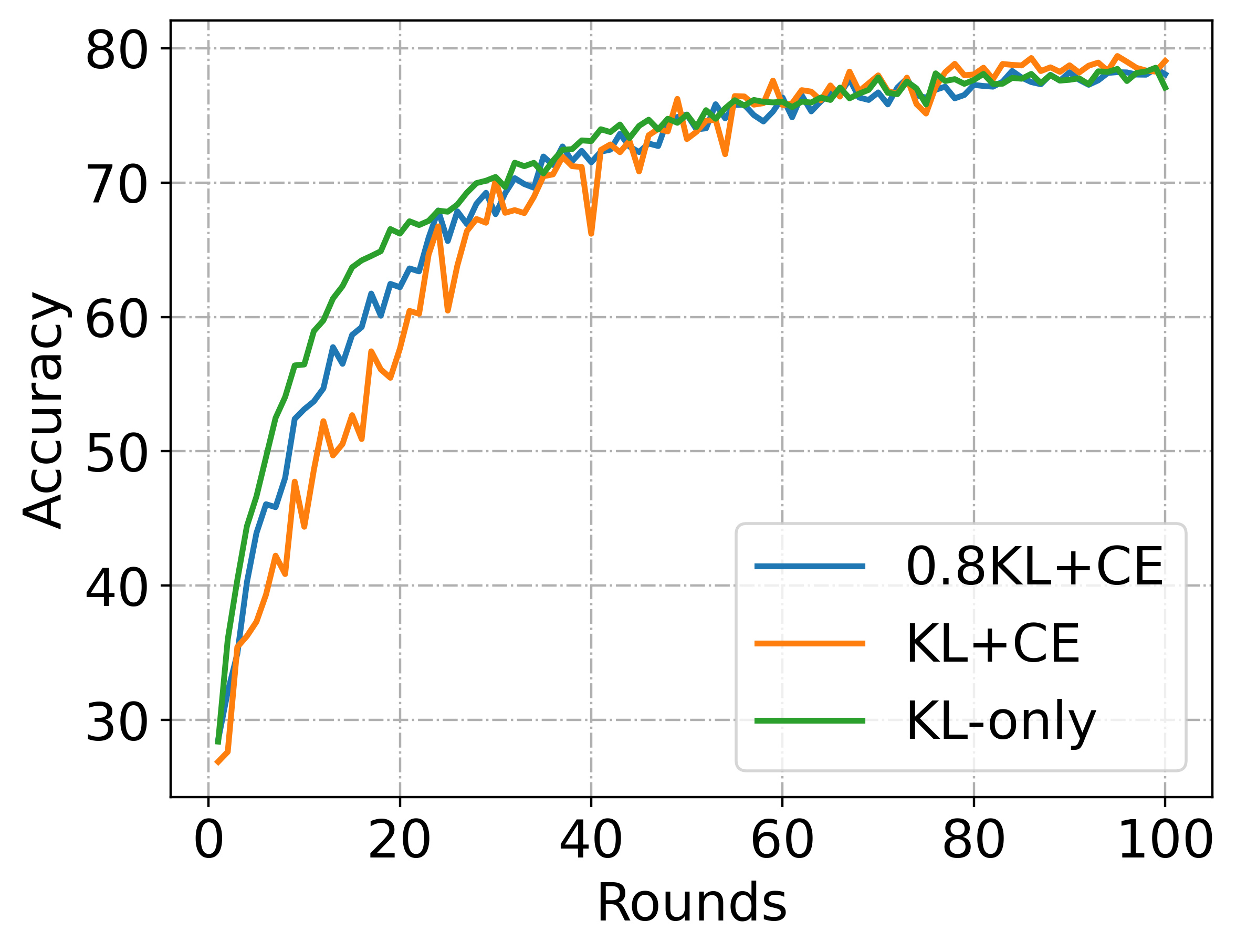}
    \caption{Test accuracy (\%) on CIFAR-10 among different loss functions in the DML process.}
    \label{img:loss}
  \end{minipage}\hfill
  \begin{minipage}{0.45\textwidth}
    \centering
    \includegraphics[width=\linewidth]{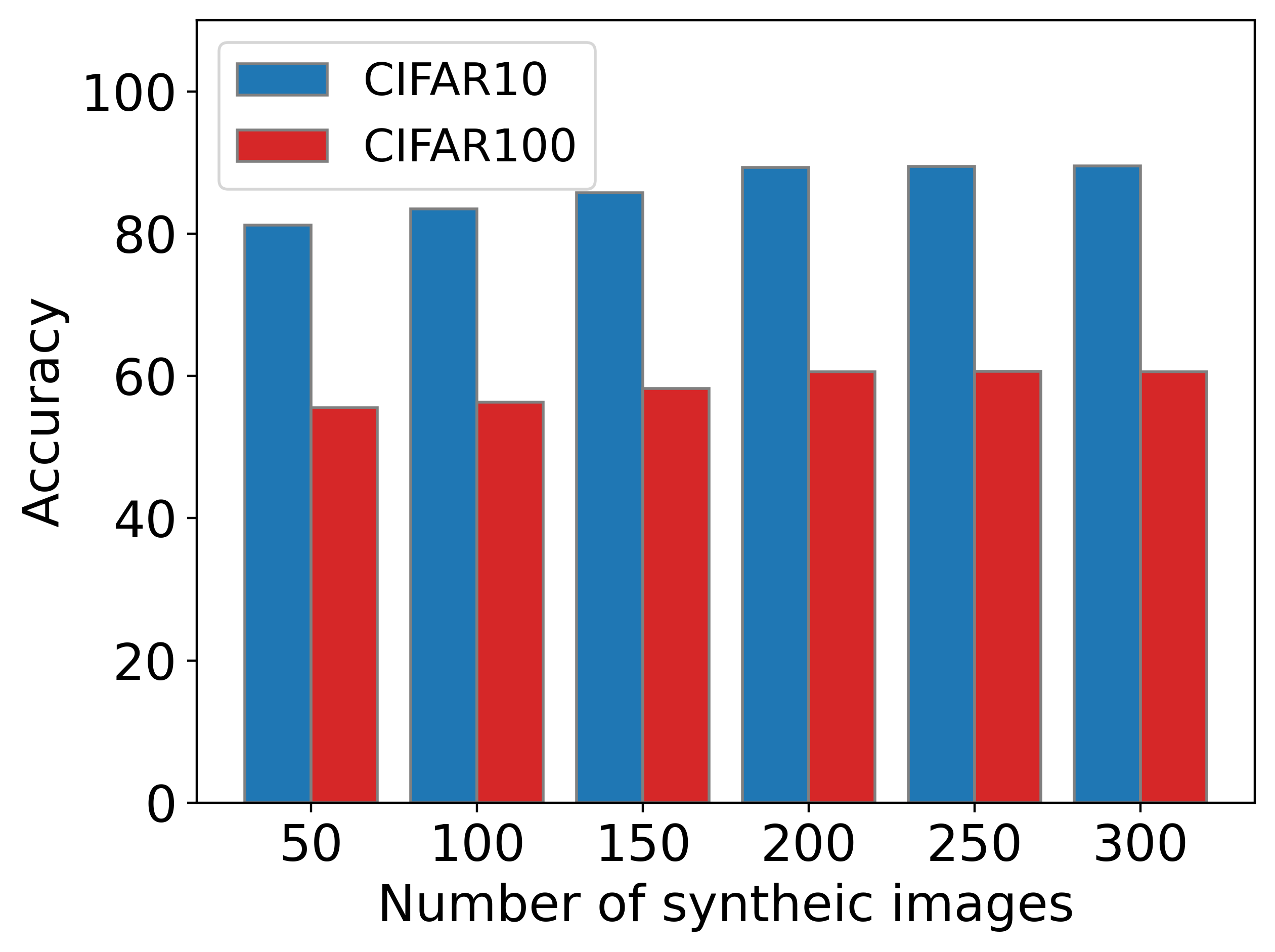}
    \caption{Test accuracy (\%) among different numbers of synthetic images.}
    \label{fig:synthetic}
  \end{minipage}
\end{figure}

\textbf{Effect of Loss Function in DML.} In the DML process, we evaluate three loss functions: KL-only, CE-only, and their linear combination. The KL-only function uses only KL divergence loss, reflecting the model's reliance on ensemble knowledge from generated data. In contrast, CE-only involves training on generated data without inter-model knowledge exchange.
The linear combination loss is formulated as follows:
\begin{equation}
L_i(w_i;\mathcal{X},\mathcal{Y})=\alpha L_{kl}+(1-\alpha) L_{ce},
\end{equation}
where $L_{kl}$ is KL divergence loss, $L_{ce}$ is cross-entropy (CE) loss, and $\alpha$ is a weighting factor. Figure~\ref{img:loss} shows the accuracy differences among these approaches. KL-only converges fastest initially but ultimately underperforms KL+CE in accuracy. While KL+CE is slower and less stable initially, it achieves the highest final accuracy. When the weight of KL divergence loss decreases, for instance, from 1.0 to 0.8, as the figure shows, the performance lies between the KL-only and KL+CE with equal weights. This suggests that combining KL and CE losses enhances global aggregation, but the slight accuracy gain from adding CE loss may not justify the extra computational cost.

\textbf{Effect of the Number of Synthetic Images.} We explore how different quantities of synthetic images from the diffusion model affect model performance. As shown in Figure~\ref{fig:synthetic}, optimal accuracy is achieved with 200 images for both CIFAR-10 and CIFAR-100 datasets. Beyond this number, additional images do not significantly enhance performance. Conversely, fewer than 150 images lead to reduced efficacy. This highlights the need for a balanced amount of synthetic data in DML: too little limits knowledge quality, while too much is not resource-efficient.

\textbf{Effect of Temperature.} The temperature $T$ is a hyperparameter in the softmax function to control the sharpness of the predicted probabilities, as Eq.~\ref{eq:logits} indicates. Higher $T$ values create a uniform distribution, while lower values lead to a peaked one. Properly tuning $T$ enhances knowledge exchange in DML, as depicted in Table~\ref{fig:temp}. Optimal performance is achieved at $T=5$ for both IID and non-IID settings, with accuracy declining at higher $T$ values. Notably, the decline in performance exhibits a more gradual trend in non-IID scenarios compared to IID. This can be explained by the smoothing effect of a higher $T$ on the logits output: even if the model lacks confidence for certain classes due to an imbalanced data distribution, the student model still benefits from the teacher's guidance. Furthermore, the smoothed logits help to prevent the student model from overfitting to classes that have more training data.

\textbf{Effect of Prompts.} We verify the impact of different prompts on the model performance and test a range of prompts from simple to descriptive, as detailed in Table~\ref{prompt}. Taking 'bird' as an example, we observe that prompt quality significantly impacts performance. A single-word prompt results in 81.31\%, while more descriptive prompts improve the accuracy. We explore the reason behind it by examining the generated data and find that simple prompts like a single word can produce abstract images, such as hand-drawn or stitching-style images. Since the model is trained with real-world photos, it cannot generate high-quality logits on these images, leading to degraded Stage 2 aggregation and thus impacting the model performance. We also notice that the model accuracy is nearly identical when using the last two prompts, which shows that FedTSA does not require detailed information about the original dataset, offering better privacy than constructing public datasets by studying raw data.

\begin{table}[ht]
\centering
\begin{minipage}{.4\linewidth}
    \centering
    \caption{Test accuracy (\%) on CIFAR-10 among different temperatures (T) in the DML process.}
    \label{fig:temp}
    \begin{tabular}{c|c|c|c}
    \hline
    T & IID & Dir(0.6) & Dir(0.3) \\ \hline
    1               & 88.59        & 87.58             & 86.96             \\ \hline
    5               & \textbf{89.28}        & \textbf{87.80}             & \textbf{87.10}             \\ \hline
    10              & 88.97        & 87.57             & 86.84             \\ \hline
    20              & 87.30        & 86.43             & 86.12             \\ \hline
    \end{tabular}
\end{minipage}%
\begin{minipage}{.6\linewidth}
    \centering
    \caption{Test accuracy (\%) with different prompts.}
    \label{prompt}
    \begin{tabular}{c|c|c}
    \hline
    Prompt                                  &  Image &  Accuracy \\ \hline
    ``bird"                                  & \adjustbox{valign=c}{\includegraphics[width=0.7cm]{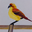}}   & 81.31          \\ \hline
    ``a real bird"                           &  \adjustbox{valign=c}{\includegraphics[width=0.7cm]{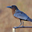}}  & 85.73          \\ \hline
    ``a photo of real bird"                  &  \adjustbox{valign=c}{\includegraphics[width=0.7cm]{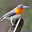}}  & \textbf{89.28}          \\ \hline
    ``a photo of real bird in CIFAR-10 style" &  \adjustbox{valign=c}{\includegraphics[width=0.7cm]{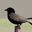}}  & 89.26          \\ \hline
    \end{tabular}
\end{minipage}
\end{table}

\section{Conclusion and Discussion}
\label{conclusion}
This paper contributes FedTSA, a novel cluster-based two-stage aggregation method for model-heterogeneous FL. FedTSA first conducts clustering to generate heterogeneous cluster-based models, then leverages DML on the server side to aggregate these models based on the data generated by a diffusion model. This approach allows each client to train a model that aligns with its available resources, thereby avoiding straggler issues and enhancing overall efficiency. Experimental results on CV datasets show that FedTSA surpasses baselines in both IID and non-IID settings. Further experiments analyze the impact of various influential factors on model training, providing a deeper understanding of their effects.

Despite the excellent performance of FedTSA, some trade-offs exist. First, the diffusion model introduces additional computational overhead to generate image data. However, it offers better flexibility in switching training tasks by simply adjusting the prompts compared to fixed public datasets. We will explore methods for achieving model-heterogeneous FL without relying on external data. Second, like other existing model-heterogeneous FL methods, FedTSA has only been verified on CV tasks. Adapting and experimenting with FedTSA on natural language processing and multimodal tasks is a primary focus for future work. 

\bibliographystyle{splncs04}
\bibliography{main}

\newpage
\appendix

\section{More Implementation Details}
\subsection{Environment}
We conduct experiments under Python 3.8.0 and PyTorch 1.31.1. We use a NVIDIA A100 provided by RunPod\footnote{https://www.runpod.io/} for computation. Weights\&Biases\footnote{https://wandb.ai/site} is leveraged to track and log the experimental results. Regarding the diffusion model, we choose the Stable Diffusion model v1-4 from Hugging Face\footnote{https://huggingface.co/CompVis/stable-diffusion-v1-4}.

\subsection{Datasets}
\textbf{CIFAR-10 and CIFAR-100.} CIFAR-10 and CIFAR-100 are benchmark datasets in the field of Computer Vision (CV), which are also popular for evaluating FL methods. CIFAR-10 consists of 60,000 32$\times$32 images in 10 classes, with 6000 images per class. CIFAR-100 also consists of 60,000 images but is categorized into 100 classes, each containing 600 images. CIFAR-10 provides a fundamental platform for image classification tasks, offering a relatively simpler challenge. CIFAR-100, on the other hand, presents a more challenging scenario with more fine-grained categories. 

\textbf{Tiny-ImageNet.} Tiny-ImageNet is a simplified version of the larger ImageNet\footnote{https://www.image-net.org/} dataset. It consists of 100,000 64$\times$64 images in 200 classes, which serves as a bridge between simpler datasets like CIFAR and the full-scale ImageNet dataset. Due to its increased complexity, training on Tiny-ImageNet is more demanding, often used to further assess a method's efficacy. It is noteworthy that accuracy levels on Tiny-ImageNet tend to be significantly lower compared to those achieved on CIFAR datasets, reflecting its heightened challenge.

\subsection{Models and the Pruned Details}
In all our experiments, we employ the ResNet18 model, which is also leveraged in baseline studies such as HeteroFL and FedProto. To introduce model heterogeneity, we apply a pruning rate to the hidden size, adjusting the input and output channels. The lower the pruning rate, the simpler the model becomes, whereas the higher the pruning rate, the more complex it is. This pruning process is applied to the first convolutional layer, all subsequent residual blocks, and the input size of the first fully connected layer.

\section{Data samples generated by diffusion model}
We implement a Conditional Generative Adversarial Network (CGAN) to generate images in the style of CIFAR-10-style and present a visualized comparison with images generated by a diffusion model. Figure~\ref{fig:ganvsdiffusion} presents the results.  In this figure, each row corresponds to a different class from CIFAR-10, showcasing 10 images per class. The class labels, arranged from top to bottom, are "airplane," "automobile," "bird," "cat," "deer," "dog," "frog," "horse," "ship," and "truck". For guiding the image generation process in the diffusion model, we employ the prompt format ``A photo of *'', where ``*'' represents the aforementioned class names.

As we can observe, it is evident that the image quality produced by the diffusion model significantly surpasses that of the CGAN-generated images. In Figure~\ref{fig:gan}, the majority only exhibit a singular outline, making the corresponding category hardly recognizable. In contrast, most images in Figure~\ref{fig:diffusion} accurately depict the corresponding class. High-quality images facilitate the generation of effective representations by models, thereby enhancing knowledge exchange efficiency and contributing to the overall improvement in model performance.

\setcounter{figure}{0}
\begin{figure}[h]
\centering
  \begin{subfigure}[b]{0.44\linewidth}
    \includegraphics[width=\linewidth]{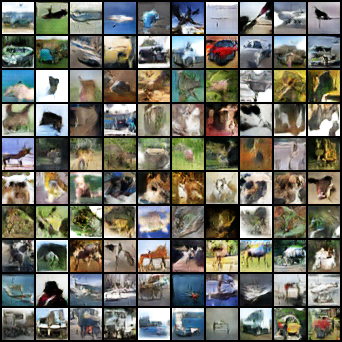}
    \caption{CGAN-generated images}
    \label{fig:gan}
  \end{subfigure}
  \hfill 
  \begin{subfigure}[b]{0.44\linewidth}
    \includegraphics[width=\linewidth]{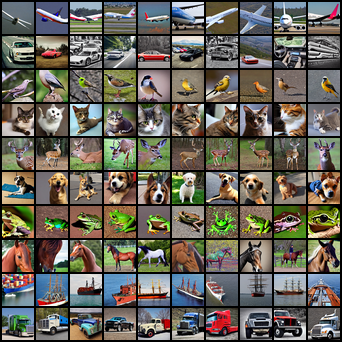}
    \caption{Diffusion model-generated images}
    \label{fig:diffusion}
  \end{subfigure}
\caption{Comparison of image generation quality: CGAN vs. Diffusion Model.}
\label{fig:ganvsdiffusion}
\end{figure}

\end{document}